\documentstyle[aaai,epsfig]{article}

\newcommand{\beq}{\begin{equation}}
\newcommand{\eeq}{\end{equation}}
\newcommand{\bdm}{\begin{displaymath}}
\newcommand{\edm}{\end{displaymath}}
\newcommand{\bea}{\begin{eqnarray}}
\newcommand{\eea}{\end{eqnarray}}
\newcommand{\nn}{\nonumber}

\newcommand{\benum}{\begin{enumerate}}
\newcommand{\eenum}{\end{enumerate}}
\newcommand{\bit}{\begin{itemize}}
\newcommand{\eit}{\end{itemize}}
\newcommand{\bdes}{\begin{description}}
\newcommand{\edes}{\end{description}}

\newcommand{\bq}{\begin{quote}}
\newcommand{\eq}{\end{quote}}

\begin{document}

%---------------------------------------

\title{Coherence, Belief Expansion and Bayesian Networks}
\author{Luc Bovens$^1$ \and Stephan Hartmann$^2$ \\
$^1$ University of Colorado at Boulder, Department of Philosophy, Boulder
CO 80309, USA \\
email:  bovens@spot.colorado.edu \\
$^2$ University of Konstanz, Department of Philosophy, 78457 Konstanz,
Germany \\
email: stephan.hartmann@uni-konstanz.de}
\maketitle

\begin{abstract}
\bq
We construct a probabilistic coherence measure for information sets which
determines
a partial coherence ordering.  This measure is applied in constructing a
criterion
for expanding our beliefs in the face of new information.  A number of
idealizations
are being made which can be relaxed by an appeal to Bayesian Networks.
\eq
\end{abstract}

%---------------------------------------

\section{Introduction}

Suppose that one receives information $\{ {\rm R_1, \dots , R_n} \}$
from $n$ independent but less than fully reliable sources.  Is it rational
to believe
this information?  Following a tradition in epistemology that goes back to John
Locke, we let belief correspond to a sufficiently high degree of confidence
(Foley
1992, Hawthorne and Bovens 1999). There are three factors that determine
this degree
of confidence: (i) How surprising is the information? (ii) How reliable are the
sources? (iii) How coherent is the information? First, suppose that the
sources are
halfway reliable and the information is halfway coherent.  Then certainly
the degree
of confidence will be greater when the reported information is less rather
than more
surprising.  Second, suppose that the information is halfway surprising and is
halfway coherent.  Let {\em truth-tellers} provide fully reliable
information and let
{\em randomizers} flip a coin for each proposition to determine whether
they will
affirm or deny it.  Then certainly the degree of confidence will be greater
when the
sources are more like truth-tellers than when they are more like
randomizers.  Third,
consider the following story: a scientist runs two independent tests to
determine the
locus of a genetic disease on the human genome.  In the first case, the tests
respectively point to two fairly narrow regions that just about overlap in
a particular region. In
the second case, the tests respectively point to fairly broad regions that
have minimal overlap in
the very same region.  Suppose that the tests are halfway reliable and that
this region is a
somewhat surprising locus for the disease.  Then certainly the degree of
confidence that the locus
of the disease is in this region is greater in the former case, in which
the information is more
coherent, than in the latter case, in which the information is less
coherent.

We define measures for each of these determinants of the degree of
confidence in a
probabilistic framework.  The real challenge lies in developing a measure of
coherence (cf. Lewis 1946, Bonjour 1985).  This measure defines a partial
ordering
over information sets.  Subsequently, we argue that belief expansion is a
function of
the reliability of the sources and the coherence of the new information
with the
information that we already believe.  We construct an acceptance measure which
determines whether newly acquired information can be added to our beliefs under
alternative suppositions about the reliability of the sources.  Our
calculations rest
on some results in the theory of Bayesian Networks.  Throughout we have
made some
strong idealizations.  We show how these idealizations can be relaxed by
directly
invoking Bayesian Networks.

%---------------------------------------

\section{The Model}

For each proposition ${\rm R_i}$ (in roman script) in the information set,
let us
define a propositional variable $R_i$  (in italic script) which can take on two
values, viz. ${\rm R_i}$ and ${\rm \overline{R}_i}$ (i.e. not-${\rm R_i}$) for
${\rm i=1,\dots , n}$.  Let $REPR_i$  be a propositional variable which can
take on two
values, viz. ${\rm REPR_i}$, i.e. there is a report from the proper source
to the
effect that ${\rm R_i}$ is true, and  ${\rm \overline{REPR}_i}$, i.e. there
is a report
to the effect that ${\rm R_i}$ is false.  We construct a joint probability
distribution
$P$ over $R_1, \dots , R_n, REPR_1,\dots , REPR_n$ satisfying the
constraint that the
sources are independent and less than fully reliable.

We model independence by stipulating that $P$ respects the following
conditional
independences:
\beq
\label{e1.1}
REPR_i \perp R_j , REPR_j \vert R_i \
{\rm for} \ i \neq j; \ i,j=1,2,\dots , n
\eeq
or, in words, $REPR_i$  is probabilistically independent of $R_j, REPR_j$,
given
$R_i$,  for $i \neq j$ and $i, j = 1, 2,\dots , n$. What this means is that the
probability that I will receive a report that ${\rm R_i}$ given that ${\rm
R_i}$ is
the case or given that ${\rm R_i}$ is not the case, is not affected by any
additional
information about whether ${\rm R_j}$ is the case or whether there is a
report to the
effect that ${\rm R_j}$ is the case.  Each source tunes in on the item of
information
that it is meant to report on: it may not always provide an accurate
report, but its
report is not affected by what other sources have to report or by other
items of
information than the one it reports on (Lewis 1946, Bovens and Olsson 1999).

We define a less-than-fully-reliable source as a source that is better than a
randomizer, but short of being a truth-teller and make the simplifying
idealization
that the information sources are equally reliable.  We specify the
following two
parameters: $P({\rm REPR_i \vert R_i}) = p$ and $P({\rm REPR_i \vert
\overline{R}_i}
) = q$ for ${\rm i=1,\dots , n}$.  If the information sources are
truth-tellers, then
$p=1$ and
$q=0$, while if they are randomizers, then $p = q > 0$. We model
less-than-full-reliability by imposing the following constraint on $P$:
\beq
\label{e1.2}
P( {\rm REPR_i \vert R_i } ) = p > q = P( {\rm REPR_i \vert \overline{R}_i
} )  > 0
\eeq

The degree of confidence in the content of the information set is the
posterior joint
probability after all the reports have come in:
\beq
\label{e1.3}
P^{\ast} ({\rm R_1,\dots , R_n}) = P( {\rm R_1,\dots , R_n  \vert
REPR_1,\dots ,
REPR_n})
\eeq

The motivation for the definition of less-than-full reliability is that we are
interested in cases in which incoming information raises our confidence in the
content of the information set to different levels.  When the sources are
randomizers, our confidence will be unaffected (Huemer 1997, Bovens and Olsson
1999), i.e. $P^{\ast}({\rm R_1,\dots , R_n}) = {\em P(} {\rm R_1,\dots ,
R_n})$; when
they are truth-tellers, our confidence will be raised to certainty, i.e.
$P^{\ast}({\rm R_1,\dots , R_n}) = 1$; and when they are worse than
randomizers, our
confidence will drop, i.e. $P^{\ast}({\rm R_1,\dots , R_n}) < {\em P(} {\rm
R_1,\dots ,
R_n})$.

%---------------------------------------

\section{Expectation, Reliability and Coherence}

It can be shown by the probability calculus, that, given the constraints on
$P$ in
(\ref{e1.1}) and (\ref{e1.2}),
\beq
\label{e2.1}
P^{\ast}({\rm R_1,\dots , R_n}) = \frac{a_0}{\sum_{i=0}^{n} a_i x^i} ,
\eeq
in which the {\em likelihood ratio} $x = q/p$ (note that $0<x<1$ for
$p>q>0$) and $a_i$ is the sum
of the joint probabilities of all combinations of values of the variables
$R_1,\dots , R_n$ that
have
$i$ negative values and
$n-i$ positive values: e.~g. for $n=3$, $a_2 = P({\rm R_1,\overline{R}_2,
\overline{R}_3 })
+ P({\rm \overline{R}_1,R_2,\overline{R}_3 }) + P({\rm
\overline{R}_1,\overline{R}_2,
R_3 })$. Note that $\sum_{i=0}^{n} a_i = 1$.

We can directly identify the first determinant of the degree of confidence
in  the
information set. Note that $a_0 = P({\rm R_1, \dots , R_n})$ is the prior joint
probability of the propositions in the information set, i.e. the
probability before
any information was received.  This prior probability is lower for more
surprising
information and higher for less surprising information.  Since more surprising
information is tantamount to less expected information, let us call this prior
probability the {\em expectation measure}.  It is easy to see that
$P^{\ast}({\rm R_1,\dots , R_n})$ is a monotonically increasing function of
$a_0$.  We
can also directly identify the second determinant, i.e. the reliability of the
sources.  Note that $P^{\ast}({\rm R_1,\dots , R_n})$ is a monotonically
decreasing
function of $x=q/p$. Hence, let us call $r:=1-x$ the {\em
reliability measure}, since $P^{\ast}({\rm R_1,\dots , R_n})$ is a
monotonically increasing function of $r$ and this measure ranges from $0$ for
sources that are randomizers to $1$ for sources that are truth-tellers.

It is more difficult to construct a {\em coherence measure}.  Consider the
following analogy: to assess the impact of a training program, we consider
the rate
of the student's actual performance level over the performance level that
he would
have reached in an ideal training program, all other things equal.
Similarly, to
assess the impact of coherence, we consider the rate of the present degree of
confidence over the degree of confidence that would have been obtained had the
information set been maximally coherent, all other things equal.  The
information set
would have been maximally coherent if and only if ${\rm R_1,\dots , R_n}$
had all
been coextensive.  Let $P$ be the actual joint probability distribution.
Construct a
joint probability distribution $P^{max}$ with the same expectation measure
and the
same reliability measure as $P$, but ${\rm R_1,\dots , R_n}$ are all
coextensive, i.e., on $P^{max}$, $a_0$ is the same as on $P$, but
$a_n = 1 - a_0 =: \overline{a}_0$, so that $a_i = 0$, for all $i \neq 0,
n$.  It follows from
(\ref{e2.1}) that,
\beq
\label{e2.2}
P^{max \ast} ({\rm R_1,\dots ,  R_n}) = \frac{a_0}{a_0 + \overline{a}_0 x^n} .
\eeq
Hence, for $a_0 \neq 0$, the ratio
\bea
\label{e2.3}
c_x({\rm R_1,\dots , R_n}) &=&
\frac{P^{\ast}({\rm R_1,\dots , R_n})}{P^{max \ast} ({\rm R_1,\dots ,
R_n})} \nn \\
&=&\frac{a_0 + \overline{a}_0 x^n}{\sum_{i=0}^{n} a_i x^i}
\eea
is a measure of the impact of the coherence of the information set on the
degree of
confidence in the content of the information set.  But note that this
measure is
contingent on the value of the reliability measure: (\ref{e2.3}) only
provides us with
a reliability-relative coherence measure.  This is unwelcome: there is a
pretheoretical notion of the coherence of an information set which has
nothing to do
with the reliability of the sources that provides us with their content.
On the
other hand, this pretheoretical notion seems to be an ordinal rather than a
cardinal
notion.  And furthermore, it seems to require a partial rather than a complete
ordering over information sets: for certain, though not for all pairs of
information
sets, we are prepared to pass a judgment that one set in the pair is more
or less
coherent than the other.

It turns out that the reliability-relative coherence measure indeed induces
a partial
ordering over informations sets which is not contingent on the reliability
of the sources.
Consider two information sets of size $n$.  These sets can be represented
by the marginal
probability distributions $P$ and $P'$ over $R_1,\dots , R_n$.  It can be
shown that for some $P$ with $\langle a_0,\dots  , a_n \rangle$ and $P'$
with $\langle
a_0',\dots ,  a_n' \rangle$, the difference $c_x({\rm R_1,\dots ,  R_n}) -
c_x'({\rm R_1,\dots  ,
R_n})$ has the same sign for any value of $x$ ranging from $0$ to $1$.
Hence, the
reliability-relative coherence measure $c_x({\rm R_1,\dots , R_n})$ induces
a partial
coherence ordering over information sets that is not contingent on the
reliability of
the sources.  For information {\em pairs}, i.e. for information sets containing
exactly two propositions, it can be shown that the following is a necessary and
sufficient condition for inclusion in the partial coherence ordering: $P$ and
$P'$ are such that (i) $a_0/a_0' \leq  a_1/a_1'$ and $a_1 \geq a_1'$, or, (ii)
$a_0/a_0' \geq a_1/a_1'$ and $a_1 \leq a_1'$.  For information sets in
general, it can
be shown that the following is a sufficient condition for inclusion in the
partial
coherence ordering: $P$ and $P'$ are such that (i) $a_i/a_i' < a_0/a_0' <
1$, or, (ii) $a_i/a_i' > a_0/a_0' > 1$, for $i=1,\dots ,  n-1$.

We provide an example of this condition for information pairs.  Suppose that
we are trying to locate a corpse of a murder somewhere in Tokyo.  We draw a
grid of
$100$ squares over the map of the city so that it is equally probable that
the murder
occurred in each grid.  We interview two independent less-than-fully-reliable
sources.  Source $1$ reports that the corpse is somewhere in squares $41$
to $60$ and
source $2$ reports that the corpse is somewhere in squares $51$ to $70$.
In this case,
$a_0=.10$ and $a_1=.20$.  This our base case. Now consider alternate case
$A$ in
which source $1$ reports squares $50$ to $60$ and source $2$ reports
squares $51$ to $61$.  In this
case, $a_0'=.10$ and $a_1'=.02$.  The information set in alternate case $A$
is clearly
more coherent than in the base case. Notice that the condition for a
partial ordering
is indeed satisfied.  But now consider alternate case $B$: source $1$
reports squares $26$ to
$60$ and source $2$ reports squares $41$ to $75$.  In this case $a_0''=.20$
and $a_1''=.30$.  Is
the information set in alternate case $B$ more coherent than in the base
case?  The
proportion of the reported squares that overlap in each report is greater
in the
alternate case, which suggests that there is more coherence.  But on the
other hand,
the price of getting more proportional overlap is that the overlapping area
is less
precise and that both sources make a much broader sweep over the map,
suggesting less
coherence.  Indeed, in this case, we cannot pass judgment whether the
information set
in alternate case $B$ is more coherent than in the base case.  Notice that the
condition for a partial ordering is indeed not justified.

%---------------------------------------

\section{Belief Expansion}

Suppose that we acquire various items of background information from
various sources
and that our degree of confidence in the content of the information set is
sufficiently high to believe the information.  Now a new item of
information is being
presented.  Are we justified to add this new item of information to what we
already
believe?  The answer to this question has something to do (i) with the
reliability of
the information source as well as (ii) with the plausibility of the new
information,
given what we already believe, or in other words, with how well the new
information
coheres with the background information.  The more reliable the source is,
the less
plausible the new information needs to be, given what we already believe, to be
justified to add the new information.  The more plausible the new
information is,
given what we already believe, the less reliable the source needs to be, to be
justified to add the new information.  The challenge is: can a precise
account of
this relationship be provided?

Our approach is markedly different from AGM belief revision.  In the AGM
approach, the question is not {\em whether} to accept new information or
not, but rather,
once we have made the decision to accept the new information, {\em how} we
should revise
our beliefs in the face of inconsistency (Makinson 1997, Olsson 1997).  Our
approach
shares a common motivation with the program of non-prioritized belief
revision.
According to Hansson (1997), we may not be willing to accept the new
information
because ``it may be less reliable (\dots) than conflicting old
information.''  Makinson
(1997) writes that ``we may not want to give top priority to new
information (\dots)
we may wish to weigh it against old material, and if it is really just too
far-fetched
or incredible, we may not wish to accept it.''  However, whereas the program of
non-prioritized belief revision operates within a logicist framework, we
construct a
probabilistic model.  The cost of this approach is that it is
informationally more
demanding. The benefit is that it is empirically more adequate, because it is
sensitive to degrees of reliability and coherence and to their interplay in
belief
acceptance.  In non-prioritized belief revision, the reliability of the
sources does
not enter into the model itself and the lack of coherence of an information
set is
understood in terms of logical inconsistency, which is only a limiting case
in our
model.  To introduce the approach, we address the question of belief
expansion.  We believe that our model also carries a promise to handle belief
revision in general, but this project is beyond the scope of this paper.

We need to make some simplifying assumptions about the origin of the background
information and the new information: (a) the propositions in the background
information are provided by independent sources, which are (b) less than fully
reliable, (c) equally reliable as the new source, and (d) independent of
the new
source.

Our background information is contained in $\{ {\rm R_1,\dots ,  R_n} \}$.
Now suppose
that we have a certain threshold level for belief and that the degree of
confidence
for the background information after having received a report to this
effect from
independent less than fully reliable sources is right at this level.  (This
stipulation is not required if we model actual cases by means of Bayesian
Networks.)
Now we are handed a new item of information ${\rm R_{n+1}}$ by an
independent less than fully
reliable source.  Then we will expand our belief set from $\{ {\rm
R_1,\dots ,  R_n}
\}$ to $\{ {\rm R_1,\dots , R_{n+1}} \}$ if and only if
\bea
\label{e3.1}
P({\rm R_1,\dots , R_{n+1}}  \vert {\rm REPR_1, \dots ,  REPR_{n+1}}) \geq
\nn \\
P({\rm R_1,\dots , R_n}  \vert {\rm REPR_1, \dots ,  REPR_n}).
\eea
Our sources are independent:
\beq
\label{e3.2}
REPR_i \perp R_j,REPR_j \vert R_i \ {\rm for} \ i \neq j; \ i,j=1,\dots , n+1
\eeq
(\ref{e2.1}) defines an {\em acceptance measure} for an information set:
\beq
\label{e3.3}
e_x ({\rm R_1,\dots ,  R_m}) =  P^{\ast} ({\rm R_1, \dots ,  R_m}) =
\frac{a_0}{\sum_{i=0}^{m} a_i x^i}
\eeq

Considering (\ref{e2.3}) and (\ref{e3.3}), we can define this acceptance
measure in terms of the reliability-relative coherence measure $c_x$,
provided that $a_0 \neq 0$:
\beq
\label{e3.4}
e_x ({\rm R_1,\dots , R_m}) = \frac{a_0}{a_0 + \overline{a}_0 x^m} c_x({\rm
R_1,\dots ,  R_m})
\eeq

>From (\ref{e3.1}) and (\ref{e3.3}), it follows that we can expand our
belief set with
a new item of information if and only if
\beq
\label{e3.5}
e_x ({\rm R_1,\dots , R_{n+1}}) \geq e_x ({\rm R_1,\dots , R_n}).
\eeq
We can make the following two observations:

\benum
\item[(i)] From (\ref{e3.3}) and (\ref{e3.5}), it is clear that whether we
can expand our
beliefs or not, is a complex function of the reliability of the sources and the
dependence of new on earlier information as expressed in the probability
distribution over the variables $R_1,\dots , R_{n+1}$.  The reliability of the
sources is reflected in the likelihood ratio $x$ and the dependence of new on
earlier information is reflected in the series $\langle a_0,\dots , a_n
\rangle$ for $e_x ({\rm
R_1,\dots , R_n})$ and in the series $\langle a_0',\dots , a_{n+1}'
\rangle$ for $e_x
({\rm R_1,\dots , R_{n+1}})$.

\item[(ii)] From (\ref{e3.4}), it is clear that the acceptance measure is a
weighted
reliability-relative coherence measure.  The weight tends to 1 for smaller
values of $x$, i.e. for
more reliable sources, and for greater values of $n$, i.e. for larger
information sets, so
that the acceptance measure will coincide with $c_x$.  We have shown that
this measure lets us construct a coherence ordering over a pair of information
$n$-tuples, if certain conditions are met.  We conjecture that such an
ordering can
also be constructed over pairs containing an information $n$-tuple and an
expansion of
this $n$-tuple, i.e. over pairs of the form $\{\{ {\rm R_1,\dots , R_n} \}, \{
{\rm R_1, \dots , R_{n+1}} \} \}$, if certain conditions are met.
Contingent on
this conjecture, we can make a substantial point: if there exists a determinate
answer to the relative coherence of the old and the new information sets,
then the
more reliable the sources are and the larger the information set is, the
more the
question of belief expansion is determined by whether the new information
set is or
is not more coherent than the old information set, and not by the
reliability of the
sources.
\eenum

The acceptance measure depends, at least to some extent, on the value of the
likelihood ratio $x$.  But what, one might ask, should we do when we have
no clue
whatsoever about the reliability of the sources, except that they are
better than mere
randomizers and yet less than fully reliable?  Let us model our limited
knowledge as
a uniform distribution over the values $p$ and $q$ under the constraint
that $p>q$.
Then we can construct the following {\em averaged acceptance measure}:
\bea
\label{e3.6}
E({\rm R_1,\dots , R_m}) &=& \int_0^1 \int_0^p e_{q/p} ({\rm R_1,\dots ,
R_m}) \  dqdp  \nn
\\ &=& \int_0^1 e_x ({\rm R_1,\dots , R_m}) \ dx
\eea
We can formulate a general criterion for belief acceptance: when we have
limited
knowledge about the reliability of our information sources, we can expand
our belief
set from $\{ {\rm R_1, \dots , R_n} \}$ to $\{ {\rm R_1, \dots ,
R_{n+1}}\}$ if and only if
\beq
\label{e3.7}
E({\rm R_1,\dots , R_{n+1}}) \geq E({\rm R_1,\dots , R_n}).
\eeq

%---------------------------------------

\section{Bayesian Networks}

Bayesian Networks represent (conditional) independences between variables and
when implemented on a computer they perform complex probabilistic
calculations at the
touch of a keystroke.  We are assuming here that the reader has some
familiarity with
Bayesian Networks (Cowell et. al. 1999, Jensen 1996, Neapolitan 1990,
Pearl 1988).

We construct a Bayesian Network that permits us to read off the
reliability-relative coherence
measure of an information set $\{ {\rm R_1,\dots , R_n}\}$ in Figure $1$.
First, we
construct a Bayesian Network with nodes for the variables $R_1,\dots , R_n$
which
represents the marginal probability distribution over these variables.
Then we add
nodes for the variables $REPR_1, \dots , REPR_n$ and draw in an arrow from
each node for the variable $R_i$  to the node for the variable $REPR_i$ and
specify
the conditional probabilities in (\ref{e1.2}) for each arrow.  By the standard
criterion of $d$-separation, we can now read off the conditional
independences in
(\ref{e1.1}) from the network.  Subsequently, we construct a node for the
variable
$R_1 \& \dots \&R_n$: we draw in the arrows and specify conditional
probabilities
such that ${\rm R_1 \& \dots  \&R_n}$ holds if and only if  ${\rm R_1,\dots
,}$ and ${\rm R_n}$
hold.  We can now read off $P^{\ast}({\rm R_1,\dots , R_n})$: it is the
probability of ${\rm R_1
\& \dots \&R_n}$ after instantiating ${\rm REPR_1,\dots , REPR_n}$.  To
read off $P^{max \ast }({\rm R_1,\dots , R_n})$, more construction is
needed.  Notice that
$P^{max} ({\rm R_i}) = P^{max} ({\rm R_1 \& \dots \& R_n})$ for ${\rm i=1,\dots
, n}$ in the counterfactual case of maximal coherence, is equal to $P({\rm
R_1 \&
\dots \& R_n})$ in the actual case where the information set may not be
maximally
coherent.  Hence $P^{max}({\rm R_1,\dots , R_n})$ is the posterior joint
probability of ${\rm R_1,\dots , R_n}$, had we been informed in the actual
case by
$n$ less than fully reliable independent sources that ${\rm R_1 \& \dots  \&
R_n}$.  So we add nodes for the variables $REP_i \& R$  (whose positive
values states
that the $i$-th source informs us that ${\rm R_1 \& \dots \& R_n}$), draw
in the proper
arrows and specify the proper conditional probabilities.  We can now read off
$P^{max \ast}({\rm R_1,\dots , R_n})$: it is the probability of ${\rm  R_1
\& \dots \& R_n}$ after
instantiating ${\rm  REP_1 \& R,\dots , REP_n \& R}$. The measure $c_x({\rm
R_1 \& \dots \&
R_n)}$ follows by (\ref{e2.3}).

\begin{figure}[t]
%\hspace{3cm}
\begin{center}
\epsfig{file=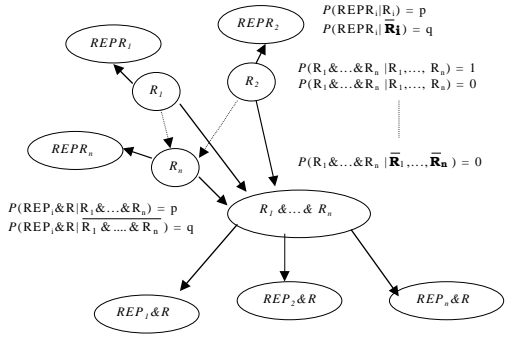,height=4.5cm}
\end{center}
%\vspace{-2cm}
\begin{center} \begin{minipage}{8cm}
\caption{Bayesian Network for coherence measure}
\end{minipage} \end{center}
%\vspace{-2cm}
\end{figure}

\begin{figure}[t]
%\hspace{3cm}
\begin{center}
\epsfig{file=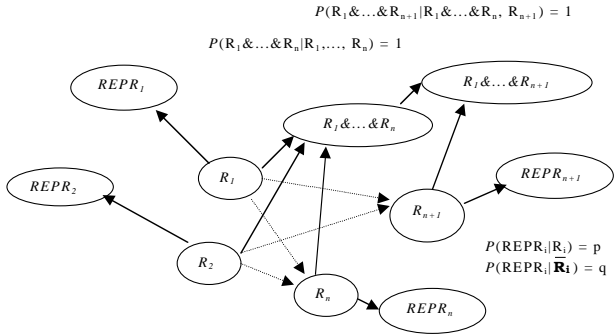,height=4.5cm}
\end{center}
%\vspace{-2cm}
\begin{center} \begin{minipage}{8cm}
\caption{Bayesian Network for belief expansion}
\end{minipage} \end{center}
%\vspace{-2cm}
\end{figure}

We construct a Bayesian Network in Figure $2$ to determine whether belief
expansion is  warranted or not. The construction of the nodes for the
variables $R_1,\dots ,
R_{n+1}$ and $REPR_1, \dots , REPR_{n+1}$ should be clear from our
construction of the Bayesian
Network in Figure $1$.  This part of the Bayesian Network respects the
conditional independences
in (\ref{e3.2}).  Now we add a node for the variable $R_1 \& \dots  \& R_n$
and a node for the
variable $R_1 \& \dots \& R_{n+1}$ and specify the conditional
probabilities so that ${\rm  R_1 \&
\dots  \& R_n}$ holds if and only if ${\rm R_1,\dots ,}$ and ${\rm R_n}$
hold and ${\rm R_1 \&
\dots \& R_{n+1}}$ holds if and only if ${\rm R_1,\dots , R_n}$ and ${\rm
R_{n+1}}$ hold.  We
instantiate ${\rm REPR_1,\dots , REPR_n}$ and propagate the evidence
throughout the network.  We
can now read off the acceptance measure $e_x({\rm R_1, \dots , R_n})$ which
is the posterior
probability of ${\rm  R_1 \& \dots  \& R_n}$.  To raise the question of
belief expansion, this
value should be greater than or equal to our threshold value for belief.
Subsequently, we instantiate ${\rm REPR_{n+1}}$ and propagate the evidence
throughout
the network.  We can now read off the acceptance measure $e_x({\rm R_1, \dots ,
R_{n+1}})$ which is the posterior probability of ${\rm R_1 \& \dots  \&
R_{n+1}}$.
Depending on our treshold value for belief, we can determine whether we are
justified
to expand our beliefs with the proposition ${\rm R_{n+1}}$.

It is easy to see how the idealizations can be relaxed in the networks.  We
can
stipulate alternative reliability parameters for the sources.  We can let
one source
report on two items of information.  We can add arrows between the $REPR_i$
variables
or between some $REPR_i$ and $R_j$ variables (for $i \neq j$) to model
certain types
of dependence between the sources. It suffices that $P({\rm R_1,\dots ,
R_{n}})$ is
equal to or greater than the threshold value for belief.  Furthermore, even if
$P({\rm R_1,\dots , R_{n+1}})$ is below the threshold value for belief, the
model
yields a marginal probability distribution over $R_1,\dots , R_n$.  Hence,
the general
question of belief revision becomes a question of defining a function which
maps  joint
probability distributions over a set of propositional variables into sets
of propositions that are
values of a subset of these variables and that can reasonably be believed.
Defining such a function
is beyond the scope of this paper.

%---------------------------------------

\section{Conclusion}

(i) We have designed a procedure to determine a partial coherence ordering
over a set
of information sets of size $n$.  If one information set is more coherent
than another
on this ordering, then our degree of confidence in the content of the
former set will
be greater than in the content of the latter set, after having been informed by
independent and less than fully reliable sources, ceteris paribus. (ii) We have
designed a probabilistic criterion for (non-prioritized) belief expansion,
which
determines whether it is rational to believe new information, considering how
reliable the sources are and how well the new information coheres with the old
information. (iii) If either the sources are sufficiently reliable or the
information
set sufficiently large, then the question of belief expansion is largely
determined by
whether the expanded information set is more coherent than the original
information
set (provided that there exists an ordering of this pair of information
sets), and only
marginally by the reliability of the sources. (iv) We have shown how a
coherence
ordering over information sets can be constructed by means of Bayesian
Networks and
how belief expansion can be modeled by means of Bayesian Networks in an
empirically
adequate manner.

%---------------------------------------

\bigskip

\parindent0em

{\bf Acknowledgments.} Thanks to Erik Olsson and the NMR referees for
discussion,
comments or suggestions. The research was supported by the
Alexander-von-Humboldt Foundation and
the German-American Council Foundation.

%---------------------------------------

\bigskip

{\bf References.}

\bdes
\item Bonjour, L. 1985. {\em The Structure of Empirical Knowledge}.
Cambridge, Mass.:
Harvard  University Press.

\item Bovens, L., and Olsson, E. 1999. Coherentism, Reliability and
Bayesian Networks.
{\em Technical Report, Logik in der Philosophie - 36}, Department of
Philosophy, University of
Konstanz.

\item Hawthorne, J., and Bovens, L. 1999. The Preface, the Lottery and the
Logic
of Belief. {\em Mind} 108: 241-264.

\item Huemer, M. 1997. Probability and Coherence Justification. {\em
Southern Journal of
Philosophy} 35: 463-472.

\item Cowell, R.~G., Dawid, A.~P., Lauritzen, S.~L., and Spiegelhafter,
D.~J. 1999.
{\em Probabilistic Networks and Expert Systems}.  New York: Springer.

\item Foley, R. 1992. The Epistemology of Belief and the Epistemology of
Degrees
of Belief. {\em American Philosophical Quarterly} 29: 111-121.

\item Jensen, F.~V. 1996. {\em An Introduction to Bayesian Networks}.
Berlin: Springer.

\item Lewis, C.~I. 1946. {\em An Analysis of Knowledge and Valuation}.
LaSalle, Ill.: Open
Court.

\item Makinson, D. 1997. Screened Revision. {\em Theoria} 63: 14-23.

\item Neapolitan, R.~E.  1990. {\em Probabilistic Reasoning in Expert
Systems}. New York:
Wiley.

\item Olsson, E. 1997. A Coherence Interpretation of Semi-Revision. {\em
Theoria} 63:105-133.

\item Pearl, J. 1988. {\em Probabilistic Reasoning in Intelligent Systems}.
San Mateo, Calif.:
Morgan Kaufmann.

\edes

\end{document}